\documentstyle[12pt]{article}

\begin{document}

\title{About compression of vocabulary 
in computer oriented languages}

\author{V.~P.~Maslov}
\date{}

\maketitle

\begin{abstract}
The author uses the entropy of the ideal Bose-Einstein gas 
to minimize losses in computer-oriented languages.
\end{abstract}

Natural languages have the property of being excessive, 
and this excessiveness is important from the viewpoint 
of reliability of information transmission.
The excessiveness manifests itself in doubling 
the grammatical or lexical tools 
in transmission of the sense of statements.
So, at lexical level, 
the synonymy in natural languages is highly developed, 
i.e., the language contains many words and word-combinations
with closely related meanings.
At grammar level, 
the meanings are doubled in grammatical forms
of different parts of the statement.
For instance, in Russian, 
the person is given by both a pronoun 
and the ending of a verb in the Present tense, 
the spatial meaning of a noun is given by both 
the case ending and a preposition, 
etc. 
The excessiveness is especially typical of natural languages 
of synthetic type where a single word
combines several morphological units 
expressing one or other meaning 
(i.e., the root, suffices, prefices, endings; 
all these parts of the word bear definite meanings). 
Such languages comprise, for instance, 
Russian, Hungarian, Baltic languages.
Usually, in languages of this type, 
the system of cases is highly developed, for instance,
there are 6 cases in Russian and 
more than~20 cases in Hungarian. 
The case endings allow one to express 
delicate spatial meanings such as: 
position in some place, 
entering some place, position near some place,
direction of motion, etc.

In languages of analytic type, 
complicated meanings are expressed by combining 
separate words with minimal changes (inflection);
this is the simplest and most transparent (motivated) 
nomination method. 
Examples of such languages are
the English language, which contains only one case, 
the possessive case, 
and the South-East Asia languages
without any inflection at all.

The property of being excessive 
is to be contrasted with the economy 
or minimality of the expression tools.
The excessiveness of a code makes the recognition more reliable. 
However, in some situations,
the economy of language tools is most important.
In particular, 
such situations appear under natural conditions 
of constructing tools for inter-ethnic communication
in districts with mixed population, i.e., 
of constructing the so-called Pidgin languages. 
The Pidgin languages are usually formed 
by significantly simplifying 
the structure of the original language.
In Pidgin languages, first of all, 
the complicated grammatical models
of the original language are significantly simplified  
and new analytic forms are developed.
The Pidgin languages are widely used 
in the South-East Asia, Oceania, Africa. 
Several Pidgin languages on the Russian basis are also known.  
For instance, in the 19th century,  
on the Russian--Chinese frontier,
a kjachta-chinese dialect was made up 
in the trade settlement ``Maimachin'' 
(bordering on Kjachta).
This dialect was used by chinese merchants 
coming to trade in Russia. 
This Pidgin language got the name of the place 
where it was born --  the ``maimachin''. 
A special manual of the ``maimachin'' language 
was printed in China.
The chinese people going to Russia 
must pass exams in the ``maimachin'' language. 
The Russian people also used this language as a means of
intercourse with chinese merchants 
(e.g., see~\cite{Cher,Neumann}).
The Russian grammar in this dialect was simplified 
to the maximal extent so that 
it approached the Chinese grammar: 
declinations, conjugations, cases and numbers, etc.
were ``abolished.''
The synthetic forms were also absent.
For instance, the verbal grammatical meanings
were expressed by using forms of the indefinite mood,
which usually coincide with the form of the imperative mood
in Russian,
and by using ``auxiliary'' verbs for expressing the Present tense.
All inflexional forms of pronouns 
were replaced by a single form. 
The dictionary was also significantly simplified, 
i.e., it included several thousands of words, 
which must roughly express all necessary meanings.

The problem to economize and minimize 
the language tools preserving the reliability 
of information transmission 
was a pressing problem in computer science.
In designing various automated systems 
working with information given in natural languages 
(information-search systems, 
systems of automated annotating and reviewing, 
systems of automated translation, etc.), 
the problem of ``compressing'' and economical coding of texts
arises first of all.
In all these cases, 
special semantic dictionaries are constructed, 
which separate words and word-combinations
close in meaning into classes.
In this case,  
delicate semantic distinctions of elements are ignored
within a class, and the name of the class becomes a new code 
of some meaning. 
In information-search languages, 
the so-called descriptors are taken to be such artificial words.

This roughening inevitably leads 
to losses in information and to additional noises. 
Therefore, the problem of optimal choice 
of the language properties, 
such as the excessiveness and the economy, 
is very important for work with computers,
which become an integral part of modern life.

If we compare a Pidgin language with a natural language, 
then we see that the Pidgin vocabulary
is a set of descriptors, i.e., of sets 
including words and forms of words of the natural language,
which are either close in meaning 
or express the same meaning.
To express this meaning in the Pidgin language,
only one word is necessary.
Since
a highly developed dictionary is not necessary 
for everyday communications
and semantic nuances can be ignored, 
the number of such words in a descriptor 
can be rather large,
of order of several hundreds and even thousands, 
and a vocabulary of many thousands, 
for instance, that of the Russian language, 
can be separated in a significantly lesser number 
of descriptors.

Similarly to the problem of universal coding, 
the language corresponding to a set of descriptors
(e.g., to the ``mai\-ma\-chin'')
is associated with  some entropy
other than the Shannon entropy.

Let us consider the simplest example.
Suppose that a ``maimachin'' word~$\alpha$ 
corresponds to a descriptor containing two words of the Russian
language:  a word~$a$ and a word~$b$. 
Now we assume that the word~$\alpha$ was used two times
in the translation of a book   
from the Russian into the ``maimachin'' language.
Which Russian words and how many times 
were replaced by the word~$\alpha$? 
The following versions are possible:

1. the word~$a$ was replaced two times;

2. the word~$b$ was replaced two times;

3. the word~$a$ was replaced once, 
and the word~$b$ was replaced once.

In the general case, 
if the  ``maimachin'' word~$\alpha$ was used $N$ times 
in the translation of the book, 
and the number of different Russian words replaced by this word
is equal to~$G$, 
then the number of all possible versions is equal to
$$
C_N^{N+G-1} =  \frac{(N+G-1)!} {N!(G-1)!}.
$$
A similar formula is encountered in problems of universal
coding~\cite{Fit}. 

Let $k$ be the number of  ``maimachin''  words  
$\alpha_1, \dots, \alpha_k$, and let the word~$\alpha_i$
be used $N_i$ times in the translation.
Then the total number of possible combinations 
is equal to 
$$
\prod^k_{i=1} \frac{(N_i+G_i-1)!} {N_i!(G_i-1)!}.
$$

Hence (by the Stirling formula), 
for $N_i \gg 1$ and $G_i \gg 1$,
the entropy is equal to   
$$
S = \ln \prod^k_{i=1} \frac{(N_i+G_i-1)!} {N_i!(G_i-1)!} \cong
\sum_{j=i}^k G_j 
\{(1+ \bar n_j) \ln (1+\bar n_j) - \bar n_j \ln \bar n_j\}
\eqno (a),
$$
where $\bar n_j = \frac{N_i}{G_i}$.

Suppose that  $M=\sum G_i$, $ M \to \infty$, 
$ N \to \infty$, and ${N}/{M} \to \rho$.

In this case, it the frequency probabilities of~$N$ particles 
are defined as ${N_i}/{N}$,
then for $p_i = \lim\frac{N_i}{N}$ and $g_i=\lim \frac{G_i}{M}$, 
we obtain the following value of the specific entropy:
$$
S = \sum_i (p_i+q_i\rho^{-1})\ln(p_i+q_i\rho^{-1}) - p_i \ln p_i -
q_i\rho^{-1}\ln q_i\rho^{-1}.
\eqno(1)
$$
This entropy coincides with the Shannon entropy 
only if $N_i \ll G_i$.

We introduce the notion of information price 
corresponding to each ``mai\-ma\-chin'' word~$\alpha_i$,  
$i=1, \dots, k$~\cite{Strat}. 
We will minimize the entropy~$S$ under the assumption 
that the general energy satisfies the relation
$E = \sum N_i  \varepsilon_i = \sum \varepsilon_i G_i\bar n_i$,
$\sum N_i=N$.  
Constructing the Lagrange function
$ F = E -\theta S +\alpha \sum G_i \bar n_i$ 
and finding its extremum, we obtain 
$$
\bar n_j = \frac {1}{(\alpha+\varepsilon_i)\beta-1}, \qquad
\beta= \frac{1}{\theta}.
\eqno (2)
$$

Hence, calculating ${\partial F}/{\partial \theta} = S_0$ 
as a function of~$\theta$ and
$\varepsilon_1,\dots,\varepsilon_k$, 
we obtain 
$$
\varepsilon_i (E_0, \bar n_1, \dots, \bar n_k, N).
\eqno (3).
$$
The argument variables are given in advance, 
the informatibility~$\varepsilon_i$
of the $i$th dictionary can be found from implicit equations
whose solvability can be proved easily.

Precisely as the Shannon entropy coincides 
with the Boltzmann entropy (e.g., see~\cite{Teorver}, p.~36), 
the entropy~$S$ considered above
coincides with the entropy of the ideal Bose-Einstein gas
and the distribution~(1) coincides with the Bose-Einstein 
distribution (\cite{Landau}, p.~184).
However, in physics,
the ``energies''~$\varepsilon_i$ are given, while here 
they must be determined as the informatibility of words 
from a dictionary still unstudied by the linguist. 

Words in natural languages are not equivalent 
in the frequency of their use.
An index of a word function in speech,
namely, its frequency of use,
depends on the semantics, 
grammatical properties, 
the contents of texts, 
and the social-professional characteristics 
of those who speak.
In an entire language, 
the frequency of use of words is not uniform.
Calculations show that 1000 of the most common words 
cover 85\% of any text. 
To fix the frequency of use of words,
various frequency dictionaries are composed 
on the basis of different text samples.
For instance, the following dictionaries are known: 
``The frequency dictionary of the modern Russian literature 
language'' (E.~Shteinfeldt, 1973), 
``The frequency dictionary of the Russian language'' (1977),
``The frequency dictionary of M.~Yu.~Lermontov's language''
(1981), etc.

In natural languages,
the syntactic words (prepositions, conjunctions, particles)
are of the highest ranks in the frequency of use.
Next, there is a rather wide stratum of  
meaningful commonly used vocabulary,
which is mainly characterized 
by the same (rather high) frequency of use in texts.
This vocabulary is usually sufficient 
for practical communication.
Therefore, 
this vocabulary forms, as a rule, 
the basis of two-language minimal dictionaries,
which are used to adopt texts for educational purposes.

The following experience of adopting the fiction texts
in Russian. 
The famous writer, M.~Evstigneev, 
specializing in writing cheap popular fiction,
told that he had ``rewritten Gogol's works 
in a different style for people.''
This was cheap popular fiction.
Later on, 
in the first half of the eighties of the 19th century,
L.~N.~Tolstoi was seriously occupied 
with the program of rewriting fiction  
in the popular language.
His follower, V.~G.~Chertkov, 
worked on this program for more than~15 years.
Adopted fiction was also published by I.~D.~Sytin
as the library called ``Intermediary''~\cite{Sytin}.

In modern practice, 
such an adopted language is required by computers. 
The problem for linguists is to adopt the literature language  
for different purposes and to create simplified Pidgin
languages.  
But to estimate and choose artificial languages
characterized by minimal losses is a problem 
for specialists in information theory.

In this paper, 
we consider a hypothetical Pidgin language, 
i.e., the next step in simplifying 
an adopted common language,
which is constructed by using the basic lexicon  
including widely used words of the same frequency rank 
(i.e., the words mainly used with the same frequency)
and semantically covering the entire semantic space. 
This is a Pidgin language, 
which can serve as a computer oriented language
and whose vocabulary is significantly simplified 
(the number of words is essentially decreased). 
However, the dictionary can further be compressed 
by excluding the words with small~$N_i\varepsilon_i$.

The assumption that elements of the original dictionary 
are used with the same frequency 
coincides, in fact, with the quantum statistical assumption  
that the eigenvalues of the $n$-particle Schr\"odinger equation
for the ideal bose-gas are in general position 
and the additional multiplicities, 
which can appear when the eigenvalues 
of the one-particle operator are commensurable, 
are not taken into account (see~\cite{Maslov}).
Moreover, the stable situation for a dictionary, 
and hence for~$N_i$, 
is equivalent to the equilibrium distribution 
of particles in the bose-gas.

The author wishes to express his deep gratitude 
to Professor L.~A.~Bos\-sa\-ly\-go
for fruitful discussions and several valuable remarks 
of both mathematical and liguistical character.

\end{document}